\documentclass{article}
\usepackage{microtype}
\usepackage{graphicx}
\usepackage{booktabs} % for professional tables
\usepackage{hyperref}

\usepackage[accepted]{icml2023}
\usepackage{amsmath}
\usepackage{amssymb}
\usepackage{mathtools}
\usepackage{amsthm}
\usepackage{subcaption}
\usepackage{tikz}
\usepackage{pgf}
\usepackage{pgfplots}
\usepackage{pgfplotstable}
\usepackage{xcolor}
\usepackage{bbding}
\usepackage{multirow}
\usetikzlibrary{shapes.geometric, arrows, matrix, patterns}
\usepgfplotslibrary{groupplots, fillbetween, external}

\newcommand*{\refne}[1]{\tikzexternaldisable\pgfplotsplotfromname{#1}\tikzexternalenable}
\usepackage[capitalize,noabbrev]{cleveref}
\theoremstyle{plain}

\theoremstyle{definition}

\theoremstyle{remark}

\usepackage[textsize=tiny]{todonotes}
\icmltitlerunning{TeacherLM: Teaching to Fish Rather Than Giving the Fish, Language Modeling Likewise}
\begin{document}
\twocolumn[
\icmltitle{TeacherLM: Teaching to Fish Rather Than Giving the Fish, \\Language Modeling Likewise}
\icmlsetsymbol{equal}{*}
\begin{icmlauthorlist}
    \icmlauthor{Nan He$^*$}{ucas}
    \icmlauthor{Hanyu Lai$^*$}{sch}
    \icmlauthor{Chenyang Zhao$^*$}{sch}
    \icmlauthor{Zirui Cheng}{sch}
    \icmlauthor{Junting Pan}{cuhk}
    \icmlauthor{Ruoyu Qin}{sch}
    \icmlauthor{Ruofan Lu}{sch}
    \icmlauthor{Rui Lu}{sch}
    \icmlauthor{Yunchen Zhang}{uestc}
    \icmlauthor{Gangming Zhao}{hku}
    \icmlauthor{Zhaohui Hou}{bupt}
    \icmlauthor{Zhiyuan Huang}{bupt}
    \icmlauthor{Shaoqing Lu}{comp}
    \icmlauthor{Ding Liang}{comp}
    \icmlauthor{Mingjie Zhan}{comp}
\end{icmlauthorlist}

\icmlaffiliation{ucas}{University of the Chinese Academy of Sciences}
\icmlaffiliation{sch}{Tsinghua University}
\icmlaffiliation{comp}{SenseTime Research}
\icmlaffiliation{cuhk}{The Chinese University of Hong Kong}
\icmlaffiliation{uestc}{University of Electronic Science and Technology of China}
\icmlaffiliation{hku}{The University of Hong Kong}
\icmlaffiliation{bupt}{Beijing University of Posts and Telecommunications}

\icmlcorrespondingauthor{Mingjie Zhan}{zmjdll@gmail.com}
\icmlkeywords{Machine Learning, ICML}
\vskip 0.3in
]
\printAffiliationsAndNotice{\icmlEqualContribution} % otherwise use the standard text.

\begin{abstract}
  Large Language Models (LLMs) exhibit impressive reasoning and data augmentation capabilities in various NLP tasks. However, what about small models? In this work, we propose TeacherLM-7.1B, capable of annotating relevant fundamentals, chain of thought, and common mistakes for most NLP samples, which makes annotation more than just an answer, thus allowing other models to learn ``why'' instead of just ``what''. The TeacherLM-7.1B model achieved a zero-shot score of 52.3 on MMLU, surpassing most models with over 100B parameters. Even more remarkable is its data augmentation ability. Based on TeacherLM-7.1B, we augmented 58 NLP datasets and taught various student models with different parameters from OPT and BLOOM series in a multi-task setting. The experimental results indicate that the data augmentation provided by TeacherLM has brought significant benefits. We will release the TeacherLM series of models and augmented datasets as open-source.

% Large Language Models (LLMs) exhibit impressive reasoning and data augmentation capabilities in various NLP tasks. However, what about small models? In this work, we propose TeacherLM-7.1B, capable of annotating relevant fundamentals, chain of thought, and common mistakes for most NLP samples, which makes annotation more than just an answer, thus allowing other models to learn ``why'' instead of just ``what''. TeacherLM-7.1B achieved a zero-shot score of 52.3 on MMLU, surpassing most models with over 100B parameters. Even more remarkable is its data augmentation ability. Based on TeacherLM-7.1B, we augmented 58 NLP datasets and taught various student models from OPT and BLOOM in a multi-task setting. The experimental results indicate that TeacherLM's data augmentation gains clear benefits. We will release the TeacherLM series of models and augmented datasets as open-source.
\end{abstract}

\input{figure/demo.tex}

\section{Introduction}
\label{sec:intro}
Large Language Models have recently revolutionized the NLP landscape \cite{brown2020language,rae2021scaling,chowdhery2022palm,hoffmann2022training,zeng2022glm,black2022gpt,wei2022emergent,taylor2022galactica}. Compared to the increase in model size, an essential thing for achieving a deeper understanding of language is to utilize data effectively. Unfortunately, most NLP datasets have simple input-output formats, inconsistent with the data seen during the pre-training of language models. As a result, directly finetuning such data is insufficient for the model to fully understand the comprehensive content of samples, inducing difficulty in fully utilizing its learning capabilities. Thus, two leading data augmentation strategies have emerged to address this issue, including task-level and instance-level approaches.

In task-level data augmentation, the aim is to map any natural language task into multiple human-readable prompted forms with diverse wording \cite{wei2021finetuned,bach2022promptsource,wang2022super}. Furthermore, combining task-level augmentation with multi-task training leads to a more comprehensive ability. Finetuning a pre-trained model on this multi-task mixture covering various tasks \cite{sanh2021multitask,chung2022scaling,muennighoff2022crosslingual,iyer2022opt} induces the language model a more substantial zero-shot and few-shot capabilities.

Though task-level augmentation has pushed the language model's generalizability to a new height, each sample represents an individual entity, limiting the augmentation, which treats the entire dataset using a unified data augmentation method. Hence a more effective approach is to augment each sample individually based on its unique characteristics, with a new stage flourishing in instance-level data augmentation. For this approach, retrieval-based pre-trained language models \cite{borgeaud2021improving,izacard2022few} utilize small models in conjunction with a massive database to introduce more relevant knowledge for each sample. Meanwhile, LLMs can use their solid zero-shot ability to augment each sample based on different prompts \cite{wang2021towards,ho2022large}. Nevertheless, these two methods bring huge usage costs, and some of the current best-performing models are not open-source.

In order to reduce the cost of data augmentation, in this work, we propose to open-source a series of small TeacherLM models, which rival human annotation and LLMs. As an adage goes, ``It is better to teach someone how to fish than to give them the fish,'' which also applies to language models. To achieve this purpose, our approach has two primary intuitions. Firstly, we believe that the real need for augmentation in a dataset lies in each sample's label, where the model should learn ``why'' instead of just remembering ``what''. Our goal is to shift the learning objectives of language models from results-oriented to process-oriented, moving away from rote memorization towards a more holistic understanding to break through current limitations imposed on language model capabilities. Secondly, language models should mimic humans' learning process by simultaneously grasping each sample's relevant fundamentals, chain of thought, and common mistakes to understand the training objectives more comprehensively.

To achieve this, we define the learning process as comprising three dimensions: fundamental, chain of thought, and common mistake. Our ultimate goal is to annotate this information for every sample in any NLP dataset. To ensure that TeacherLM has strong zero-shot capabilities for most natural language processing tasks, we have collected 2 million samples from multiple domains for training. Furthermore, we combined manual annotation and STaR \cite{zelikman2022star} strategy to construct a complete ``\textit{\{Question\}\text{ }\{Answer\}\text{ }\{Fundamentals\}\text{ }\{Chain of Thought\}\text{ }\{Common Mistakes\}}" five-element training object for each sample.

In sum, our key contributions are:

\begin{itemize}
\item \textbf{Comprehensive}
TeacherLM can generate fundamentals, chain of thought, and common mistakes, providing comprehensive information tailored to the task's characteristics and allowing each task to learn the most relevant knowledge.

\item \textbf{Generalizability}
TeacherLM is suitable for a 
% [wide]
variety of datasets and models. As shown in Figure \ref{main_fig}, we augmented 58 NLP datasets and taught various student models with different parameters from OPT \cite{zhang2022opt}, and BLOOM \cite{scao2022bloom} series in a multi-task setting. Compared to non-augmented versions, the experimental results indicate that TeacherLM's data augmentation gains clear benefits.

\item \textbf{Fight big with small}
TeacherLM-7.1B model achieved a zero-shot score of 52.3 on MMLU, surpassing most models with over 100B parameters.

\item \textbf{Cost friendly}
TeacherLM-7.1B has only 7.1 billion parameters; compared to models such as text-davinci-003 
% [xxx] 引用
, it has efficient inference speed and lower running configuration requirements. Therefore, with the significant cost reduction, we can augment NLP datasets of millions of levels, further opening the door to the reasoning world.

\item \textbf{Open source}
We will release the TeacherLM series of models and augmented datasets as open-source.
\end{itemize}

\section{Related Work}
\label{sec:related work}
This paper explores the intersection of various NLP research fields, including multi-task learning, instruction tuning, multi-step reasoning, and data augmentation. In this section, we will introduce several key related works.

\textbf{Reasoning via finetuning}
In this study, we construct a large-scale Reasoning dataset for training TeacherLM. Previous works have utilized manually annotated multi-step reasoning to improve model performance \cite{ling2017program,camburu2018snli,rajani2019explain,hu2024minicpm,talmor2020leap,cobbe2021training,nye2021show,zelikman2022star,chung2022scaling}. Compared to these works, TeacherLM-7.1B has certain advantages when compared to models of the same scale.

\textbf{Using large models as zero-Shot data augmentation generators}
Combining chain of thought and in-context learning has unlocked more robust reasoning capabilities for 
 LLMs \cite{wei2022chain,viswanathan2023prompt2model,suzgun2022challenging,lampinen2022can}, guiding models to move from learning to rote information to now learning to think critically. Furthermore, by adding the sentence ``Let's think step by step'' before LLMs generate answers, the model can generate a step-by-step thought process and significantly improve accuracy in solving reasoning tasks\cite{kojima2022large}. Therefore, Large Language Models are zero-shot Reasoners and can also be considered as Zero-Shot data augmentation generators.

\textbf{Instruction finetuning}
Multi-task learning improves the performance of language models in zero-shot settings. Many works have found that designing elaborated natural language templates with instructions for each NLP task and connecting them breaks down barriers between tasks and allows the language model to understand the data better \cite{wei2021finetuned,sanh2021multitask,ouyang2022training,chen2024internet,wang2022super,scialom2022continual,chung2022scaling,muennighoff2022crosslingual,iyer2022opt}. In our work, we combine instruction finetuning and reasoning to unlock more potential in language models.

\section{Training TeacherLM}
\label{sec:training teacherlm}
\definecolor{tlm-560M}{HTML}{FFC3A1}
\definecolor{tlm-1.1B}{HTML}{F0997D}
\definecolor{tlm-7.1B}{HTML}{D3756B}
\definecolor{tlm-176B}{HTML}{DC0000}

\definecolor{flan-palm-8B}{HTML}{1C82AD}
\definecolor{flan-palm-62B}{HTML}{00337C}
\definecolor{flan-palm-540B}{HTML}{13005A}

\definecolor{gopher-1.4B}{HTML}{E3D18A}
\definecolor{gopher-7.1B}{HTML}{BD9354}
\definecolor{gopher-280B}{HTML}{9E7540}

\definecolor{GPT3-13B}{HTML}{86E5FF}
\definecolor{GPT3-175B}{HTML}{8BCDCD}

\definecolor{atlas}{HTML}{B3FFAE}
\definecolor{GPT-NEOX}{HTML}{82CD47}
\definecolor{chinchilla}{HTML}{54B435}
\definecolor{gal}{HTML}{379237}
\definecolor{glm}{HTML}{52734D}
\definecolor{Bloom-176B}{HTML}{007965}
\begin{figure*}[ht]
  \centering
  \begin{subfigure}{1\textwidth}\footnotesize
    \centering
      \begin{tikzpicture}
          \begin{axis}[width=1\textwidth, 
            height=0.4\textwidth,
            xmode=log,
            log ticks with fixed point,
            xmin=0.5, xmax=640, 
            xtick={0.5, 2, 10, 40, 160, 640}, 
            xlabel style={},
            xlabel={Parameters (Billion)}, 
            ylabel={Accuracy (\%)}, 
            grid=both,
            grid style={draw=gray!10},
            set layers,
            mark layer=axis tick labels
            ]

              %FLAN-PaLM
              \addplot[mark=*, color=flan-palm-540B, mark size=3] coordinates {(540, 75.2)};\label{flan-palm-540B}
              \addplot[mark=*, color=flan-palm-62B, mark size=3] coordinates {(62, 59.6)};\label{flan-palm-62B}
              \addplot[mark=*, color=flan-palm-8B, mark size=3] coordinates {(8, 49.3)};\label{flan-palm-8B}
            
              \addplot[mark=triangle*, color=chinchilla, mark size=3] coordinates {(70,67.5)};\label{chinchilla}
              %\addplot[mark=oplus*, color=teal!40, mark size=3] coordinates {(11, 48.9)};\label{unifiedQA}
              \addplot[mark=triangle*, color=gal, mark size=3] coordinates {(120, 52.6)};\label{gal}
              \addplot[mark=triangle*, color=glm, mark size=3] coordinates {(130, 45.7)};\label{glm}
              \addplot[mark=triangle*, color=atlas, mark size=3] coordinates {(11, 47.9)};\label{atlas}
              \addplot[mark=triangle*, color=Bloom-176B, mark size=3] coordinates {(176, 31.9)};\label{bloom-176B}
              \addplot[mark=triangle*, color=GPT-NEOX, mark size=3] coordinates {(20, 33.6)};\label{gpt-neox}

              %gopher
              \addplot[mark=diamond*, color=gopher-280B, mark size=3] coordinates {(280, 60.0)};\label{gopher-280B}
              \addplot[mark=diamond*, color=gopher-7.1B, mark size=3] coordinates {(7.1, 29.5)};\label{gopher-7.1B}
              \addplot[mark=diamond*, color=gopher-1.4B, mark size=3] coordinates {(1.4, 27.3)};\label{gopher-1.4B}

              %GPT
              \addplot[mark=square*, color=GPT3-175B, mark size=3] coordinates {(175, 43.9)};\label{gpt3-175B}
              \addplot[mark=square*, color=GPT3-13B, mark size=3] coordinates {(13, 26)};\label{gpt3-13B}
              %\addplot[mark=square*, color=blue!40, mark size=3] coordinates {(6.7, 43.2)};\label{gpt3-6.7B}
              %\addplot[mark=square*, color=blue!40, mark size=3] coordinates {(6, 27.3)};\label{gpt-j}
      
              \addplot[mark=none, color=teal, dashed, line width=1.5, domain=0.4:640] {89.8} node [anchor=north, color=black, xshift=-60] {Average Human Expert};
              \addplot[mark=none, color=teal!50, dashed, line width=1.5, domain=0.4:640] {34.5} node [anchor=north, color=black,  xshift=-60] {Average Human Rater};

              %TLM
              \addplot[mark=pentagon*, color=tlm-560M, mark size=4] coordinates {(0.56, 38.39)} node [anchor=north, color=black, yshift=-3] {ours};\label{tlm-560M}
              \addplot[mark=pentagon*, color=tlm-1.1B, mark size=4] coordinates {(1.1, 41.38)} node [anchor=north, color=black, yshift=-3] {ours};\label{tlm-1.1B}
              \addplot[mark=pentagon*, color=orange, mark size=4] coordinates {(3, 46.74)} node [anchor=north, color=black, yshift=-3] {ours};\label{tlm-3B}
              \addplot[mark=pentagon*, color=red, mark size=4] coordinates {(7.1, 52.3)} node [anchor=north, color=black, yshift=-3] {ours};\label{tlm-7.1B}
              \addplot[mark=pentagon*, color=tlm-176B, mark size=4] coordinates {(176, 59.8)} node [anchor=north, color=black, yshift=-3] {ours};\label{tlm-176B}
          \end{axis}
        \end{tikzpicture} 
  \end{subfigure}
  \begin{subfigure}{0.95\textwidth}
      \centering
      \begin{tikzpicture}\footnotesize
        \node(center)[draw=none]{};
        \matrix[matrix of nodes,
        draw,
        inner sep=0.005\textwidth,
        xshift=0.5\textwidth,
        ampersand replacement=\&]
        { \refne{tlm-560M}\& TeacherLM-560M \quad \& \quad
        \refne{tlm-1.1B}\& TeacherLM-1.1B \quad \& \quad
        \refne{tlm-3B}\& TeacherLM-3B \quad \& \quad
        \refne{tlm-7.1B}\& TeacherLM-7.1B \quad \& \quad \\
        \refne{tlm-176B}\& TeacherLM-176B \quad \& \quad 
        \refne{flan-palm-8B}\& Flan-PaLM-8B \quad \& \quad 
        \refne{flan-palm-62B}\& Flan-PaLM-62B \quad \& \quad
        \refne{flan-palm-540B}\& Flan-PaLM-540B \quad \& \quad \\
        \refne{gopher-1.4B}\& Gopher-1.4B \quad \& \quad 
        \refne{gopher-7.1B}\& Gopher-7.1B \quad \& \quad
        \refne{gopher-280B}\& Gopher-280B \quad \& \quad
        %\refne{gpt3-6.7B}\& GPT-3-6.7B \quad \& \quad 
        \refne{gpt3-13B}\& GPT-3-13B \quad \& \quad \\
        \refne{gpt3-175B}\& GPT-3-175B \quad \& \quad 
        \refne{atlas}\& Atlas-11B \quad \& \quad 
        %\refne{unifiedQA}\& UnifiedQA-11B \quad \& \quad 
        \refne{gpt-neox}\& GPT-Neox-20B \quad \& \quad 
        \refne{chinchilla}\& Chinchilla-70B \quad \& \quad \\
        \refne{gal}\& GAL-120B \quad \& \quad 
        \refne{glm}\& GLM-130B \quad \& \quad 
        \refne{bloom-176B}\& BLOOM-176B \quad \& \quad \\};
        %\refne{gpt-j}\& GPT-J-6B \\};
      \end{tikzpicture}
  \end{subfigure}
\caption{Average MMLU scores (\%) for 57 tasks with model and human accuracy comparisons. \textbf{TeacherLMs are in the 0-shot setting, and the rest are in the 5-shot setting.}} 
\label{mmlu-figure}
\end{figure*}
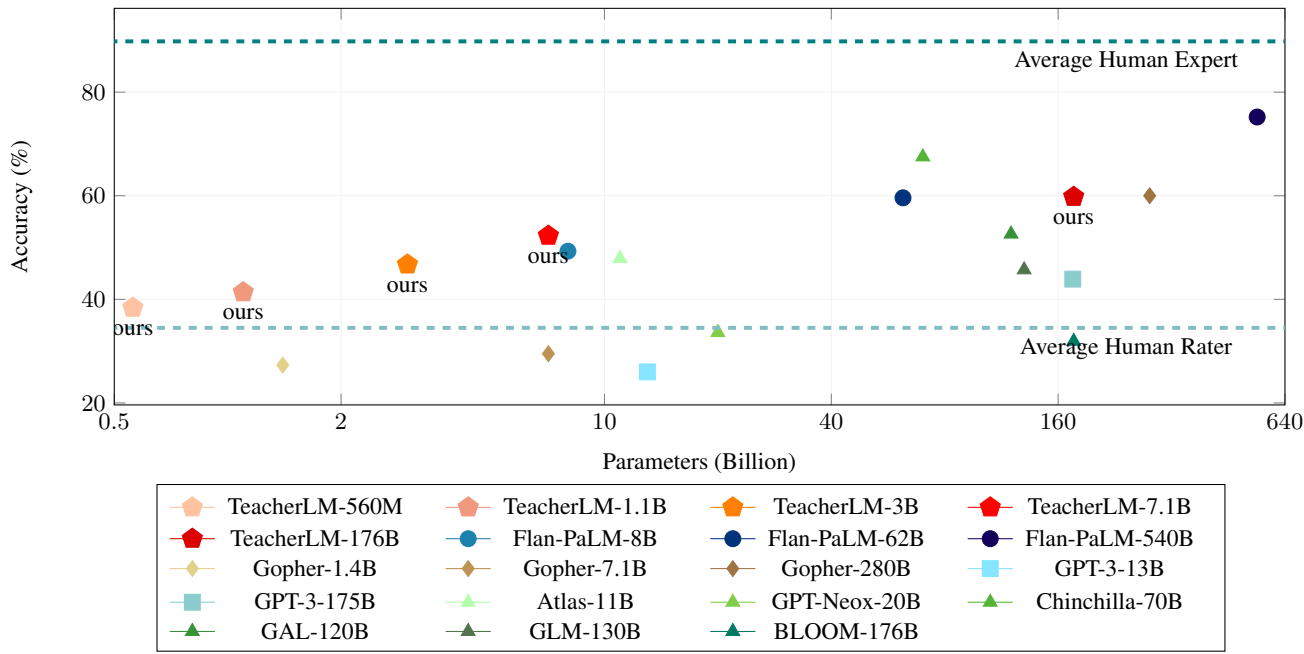
% 20230216 更改 Table 1 位置
% \input{table/TeacherLM_multistage.tex}

A good teacher can be a beacon, guiding students toward mastering the methods to solve problems. We aim the same for the TeacherLM. In this regard, we make two primary efforts. Firstly, we construct a dataset comprising two million detailed explanations. Secondly, we adopt a multi-stage progressive training mode, moving from generality towards specialization.

\subsection{Dataset Construction}

\textbf{P3-Sense-3K}
We extract 58 supervised datasets from P3 \cite{sanh2021multitask}. Each dataset contains multiple prompts, resulting in 529 tasks. To ensure sample balance, We select at most 3,000 samples of less than 1,200 tokens for each task. There are 1,400,364 samples in total. Then we format multiple choice tasks in the form of ``\textit{Q: \{question\} \{options\} A: \{answer\}}''. All other tasks are changed to the form of ``\textit{Q: \{question\} A: \{answer\}}''.

\textbf{Muffin-3W}
We extract 56 supervised datasets from Muffin \cite{wei2021finetuned}, and each dataset includes ten prompts. Similarly, We select at most 30,000 samples of less than 1,200 tokens for each task. There are 1,155,767 samples in total. All tasks are changed to the form of ``\textit{Q: \{question\} A: \{answer\}}''.

\textbf{TeacherData-2M}
Furthermore, we collect 2 million pieces of multi-domain data not included in any public datasets for training. We utilize manual annotation and the STaR \cite{zelikman2022star} strategy to construct a five-element fixed paradigm for each sample, including the question, answer, fundamentals, chain of thought, and common mistakes. We select samples of less than 2048 tokens.
 
\subsection{Training Procedure}
Training the teacher models aims to obtain checkpoints excelling at generating comprehensive explanations with solid generalization ability.
In this section, we introduce our base models and the details of the multi-stage training.

\subsubsection{Models}
In this work, our base models are the BLOOM \cite{scao2022bloom} series ranging from 560 million to 176 billion parameters, which are pre-trained on the ROOTS \cite{laurenccon2022bigscience} corpus in 46 natural languages and 13 programming languages. BLOOM models are large decoder-only language models pre-trained for 350 billion tokens.

\subsubsection{Multi-stage training}
For each model, we adopt a multi-stage training procedure. 

\textbf{Multi-task training}
Previous literature has shown that training on a large number of tasks with instructions can improve the model's zero-shot ability and allow it to generalize well on unseen tasks \cite{sanh2021multitask, wei2021finetuned}. Therefore, we combine the mixture of P3-Sense-3K and Muffin-3W to conduct multi-task training.

\textbf{Personalization training}
Using the TeacherData-2M dataset, the model simultaneously learns to analyze each sample's fundamentals, chain of thought, and common mistakes. Through this process, the models can generate three types of explanations.  

\textbf{Specialization learning}
We split the TeacherData-2M dataset and train three independent models to focus on learning fundamentals, chain of thought, and common mistakes, respectively. The resulting models are referred to as TeacherLM-Fundamental, TeacherLM-COT, and TeacherLM-CommonMistake.

In each stage of training, we use packing \cite{raffel2020exploring} to combine multiple texts, and terminators separate different texts. The interaction between texts is eliminated by setting the attention mask and resetting the position id. Models of different sizes use different learning rates and batchsizes. According to the number of parameters from small to large, they are 3e-4/256(TeacherLM-560M), 1e-4/512(TeacherLM-1.1B), 4e-5/512(TeacherLM-3B), 2e-5/768(TeacherLM-7.1B), 6e-5/1024(TeacherLM-176B). More information can be seen in Appendix \ref{appendix:hyperparameters}. After the first stage, we evaluate the models' performance on the MMLU \cite{hendrycks2020measuring} benchmark for selecting checkpoints for further training. Then, after the second and third stages, apart from evaluating performance on the MMLU benchmark, we conduct a manual evaluation as a reference for selecting checkpoints.
% 20230216 更改 Table 1 位置
\definecolor{green}{HTML}{379237}
\begin{table*}[htbp]
\caption{Average MMLU scores(\%) for 57 tasks with multi-stage training and no-stage training comparisons. C, F, M represent CoT, fundamentals, common mistakes respectively. No stage represents directly mixing all datasets for training. TeacherLM-176B only completed part of the training process and only trained on TeacherData-2M.}
\vskip 0.15in
\centering
\begin{tabular}{lcccccc}
\toprule
parameters & 1st stage & 2nd stage & 3rd stage(C) & 3rd stage(F) & 3rd stage(M) & No stage \\
\midrule
560M & 31.00 &36.47 &\underline{\textbf{\textcolor{green}{38.39}}}& 34.65 & 34.85 & 35.27 \\
1.1B & 32.87 & 41.36 & \underline{\textbf{\textcolor{green}{41.38}}} & 39.25  &38.75 &40.26 \\
3B & 36.51 & 46.34 & \underline{\textbf{\textcolor{green}{46.74}}} & 45.22 & 45.1&46.01\\
7.1B & 41.47 &51.11& \underline{\textbf{\textcolor{green}{52.30}}}  & 50.32 &49.00 &47.00 \\   
176B & /& \underline{\textbf{\textcolor{green}{59.80}}}& /& /& /&/\\
%\bottomrule
%parameters & 3rd stage(F) & 3rd stage(M) & no stage \\
%\midrule
%560M & 34.65 & 34.85 & 35.27 \\
%1.1B & 39.25  &38.75 &40.26   \\
%3B & 45.22 & 45.1&46.01 \\
%7.1B & 50.32 &49 &47  \\   
\bottomrule
\end{tabular}
\vskip -0.1in
\label{multi-stage-result}
\end{table*}
\input{figure/comparison.tex}
\subsection{Evaluation Protocol}
\subsubsection{Benchmark}
For the teacher models, we focus on the performance on the held-out tasks not included in the training procedure.
Specifically, we evaluate the teacher models on the MMLU(57 tasks)
benchmark in the zero-shot setting. The benchmark consists of exam-like questions on academic subjects such as mathematics, history, law, and medicine.

\subsubsection{Evaluation Method}
The evaluation adopts the method of rank classification. In more detail, assume answers are from a fixed set $\mathbb{A}$, and $\text{answer}_i\in\mathbb{A}$, where $i=1,...,n$, $n$ indexes $n$
% all
candidate answers. Calculate the probability of each candidate answer using the Formula: 
\begin{equation}
     P(\text{answer}_i|q)=\frac{1}{K}\sum_{k=1}^{K}\log P(a_k|q,a_1,...,a_{k-1})
\label{eval}
\end{equation}
Here $\log P(a_k|q,a_1,...,a_{k-1})$ is the log probability of generating the k-th token $a_k$ in $\text{answer}_i$ conditioned on the previous tokens. $K$ is the total number of tokens in $\text{answer}_i$ and $q$ is the question. Choose the answer with the highest probability and calculate average accuracy on all tasks.

\subsection{Results}
\subsubsection{Zero-shot Setting}
We show the MMLU scores with model and human comparisons in Figure \ref{mmlu-figure}. There are some key points.

First, models with few parameters can still benefit from our datasets. Even our 560M model exceeds BLOOM-176B. Moreover, scaling the CoT data to a larger size can improve zero-shot performance. In FLAN-PaLM \cite{chung2022scaling}, nine CoT datasets are added to improve the performance on the held-out tasks. Further scaling the number of CoT samples to 2 million in this work shows fantastic improvement.
The TeacherLM-7.1B model has achieved a zero-shot score of 52.3 on the MMLU, surpassing the 5-shot performance of most hundred billion parameter models. 
Additionally, the zero-shot score of TeacherLM-176B is 59.8, comparable to the 5-shot score of gopher-280B \cite{rae2021scaling}. The full results are in Appendix \ref{appendix:result}.

\subsubsection{Multi-stage Training Helps A Lot}
We show the evaluation results during multi-stage training in Table \ref{multi-stage-result}. In contrast to multi-stage training, we also mixed all datasets and trained on them directly. The training hyperparameters and the number of steps are consistent. We find that directly blending all datasets scores much lower than multi-stage training. However, only the models using CoT data obtained continuous improvement in the third stage. This phenomenon may indicate that tasks of the MMLU benchmark require higher reasoning ability. 

\subsubsection{Add Subject To Prompt}
Inspired by the fact that the MMLU benchmark is composed of different subject tasks, we added the prompt ``\textit{The following are multiple choice questions (with answers) about \{subject name\}}" at the beginning of each text. We labeled the subject of each sample of TeacherData-2M. This approach increases the score of TeacherLM-7.1B by 2\%. We think such a prompt is more helpful for the model to use the correct knowledge to answer the question.

\section{Teaching Student}
\label{sec:teaching student}
\begin{table*}[t]
\caption{
Dataset settings in section \ref{sec:teaching student}. Here shows option amounts, example amounts, and average tokens in the training split of each dataset. \textbf{Manual} denotes human annotated rationales in original datasets. \textbf{CoT}, \textbf{Fud}, and \textbf{Mis} are TeacherLM's generated CoT, fundamentals, and common mistakes, while \textbf{CoT-D}, \textbf{Fud-D}, and \textbf{Mis-D} are text-davinci-003's generated CoT, fundamentals and common mistakes, through the same prompts inputted to TeacherLM.
}
\label{tab:dataset}
\vskip 0.15in
\begin{center}
\begin{small}
\begin{sc}
\begin{tabular}{lcccccccccccr}
\toprule
\multirow{2}{*}{Dataset} & \multirow{2}{*}{Options} & \multicolumn{2}{c}{Exmaple Amounts} & \multicolumn{7}{c}{Average Tokens}                                                      \\ \cline{3-11}
                         & & Train             & Test            & Manual & CoT   & Fud & Mis & CoT-D & Fud-D & Mis-D \\
\midrule
ECQA                     &5 & 7598              & 2194            & 59  & 75 & 186      & 53 & 47       & 27              & 32         \\
StrategyQA               &2 & 1832              & 228             & 28  & 90 & 192      & 55 & 73       & 48              & 31         \\
CREAK                    &2 & 10174             & 1371            & 15  & 71 & 190      & 48 & 42       & 25              & 19         \\
P3-sense-3K                   &/ &1400364            &/               &/   &49  &135       & 54  &/        &/               &/          \\
\bottomrule
\end{tabular}
\end{sc}
\end{small}
\end{center}
\vskip -0.1in
\end{table*}

\definecolor{blue}{HTML}{00337C}
\definecolor{green}{HTML}{379237}
\begin{table*}[t]
\caption{Single-task finetuning results for the BLOOMZ-7.1B model and the P3-Augmented-BLOOMZ-7.1B model. Text means no data augmentation for the corresponding task, and manual means using manual CoT. Each result of TeacherLM-7.1B and text-davinci-003, from left to right, represents the score after data augmentation using the corresponding model to generate CoT, fundamentals, and common mistakes.}
\vskip 0.15in
\begin{center}
\begin{sc}
\begin{tabular}{lllll}
\toprule
task & text & manual & TeacherLM-7.1B & text-davinci-003 \\
\midrule
% &&BLOOMZ-7.1B\\
\multicolumn{5}{c}{BLOOMZ-7.1B} \\
\bottomrule
ECQA        & \underline{\textbf{\textcolor{green}{71.9}}} & 61.9 & 61.9 / 61.9 / 61.9 & 64.0 / 64.0 / 64.4 \\
StrategyQA  & 71.1 & 71.1 & 65.8 / 73.3  / 57.0 \textbf{\textcolor{blue}{(+2.2)}} & 66.7 / \underline{\textbf{\textcolor{green}{79.0}}}  / 69.3 \textbf{\textcolor{blue}{(+7.9)}}\\
CREAK       & 64.0 & 73.3 \textbf{\textcolor{blue}{(+9.3)}} & \underline{\textbf{\textcolor{green}{77.0}}}  / 54.8 / 58.2 \textbf{\textcolor{blue}{(+13.0)}} & 74.6  / 69.6  / 76.7 \textbf{\textcolor{blue}{(+12.7)}}  \\
\midrule
% &&P3-Augmented-BLOOMZ-7.1B\\
\multicolumn{5}{c}{P3-Augmented-BLOOMZ-7.1B} \\
\midrule
ECQA        & \underline{\textbf{\textcolor{green}{68.6}}} & 53.1  & 60.5 / 58.3 / 65.0  & 60.1 / 67.7 / 56.6 \\
StrategyQA  & 72.4 & 68.9 & 74.1  / \underline{\textbf{\textcolor{green}{77.2}}}  / 74.1 \textbf{\textcolor{blue}{(+4.8)}}  & 65.8 / 75.4  / 71.9 \textbf{\textcolor{blue}{(+3.0)}} \\
CREAK       & 76.7 & 78.6 \textbf{\textcolor{blue}{(+1.9)}}  & 75.9 / 75.8 / \underline{\textbf{\textcolor{green}{79.0}}} \textbf{\textcolor{blue}{(+2.3)}}  & 70.9 / 70.0 / 70.3 \\
\bottomrule
\label{tab:bloomz}
\end{tabular}
\end{sc}
\end{center}
\vskip -0.1in
\label{tab:bloomz}
\end{table*}
\definecolor{light-green}{HTML}{99E2B4}
\definecolor{light-blue}{HTML}{CAF0F8}
\definecolor{light-gray}{HTML}{EDEDE9}
\definecolor{grey}{HTML}{D6CCC2}

\begin{figure*}[!ht]
    \begin{subfigure}{0.74\textwidth}\footnotesize
        \begin{tikzpicture}

            \node(bounding box)[rectangle, minimum width=1\textwidth, minimum height=0\textheight, draw=none, dashed]{};

            \node(Input Text)[rectangle, minimum width=1.1\textwidth, minimum height=0.04\textheight, fill=grey!10, xshift=0\textwidth, text width=1.35\textwidth, align=left, yshift=0.09\textheight]{\textbf{Question:} Fact 1: Climate is generally described in terms of temperature and moisture. Fact 2: Humidity is a measure of moisture content of air. Climate is generally described in terms of temperature and what? \\ (A) humidity (B) wind (C) warmer air (D) calibrated (E) percentages (F) matter (G) color (H) SI units \\ \textbf{Answer:} The answer is (A).};

            \node(Coordinate)[rectangle, rounded corners=0.0\textwidth, minimum width=0.0\textwidth, minimum height=0.0\textheight, below of=input box, yshift=-0.025\textheight]{};

            %! 这是全局的坐标，不显示任何字样

            \node(Manual Text)[rectangle, minimum width=1.1\textwidth, minimum height=0.04\textheight, fill=grey!15, xshift=0\textwidth, text width=1.35\textwidth, align=left, below of=Coordinate, yshift=+0.0\textheight]{\textbf{Manual Explanations}\\Climate is generally described in terms of temperature and humidity.};

            \node(Davinci Chain-of-Thought Text)[rectangle, minimum width=1.1\textwidth, minimum height=0.04\textheight, fill=light-blue!30, xshift=0\textwidth, text width=1.35\textwidth, align=left, below of=Coordinate, yshift=-0.04\textheight]{\textbf{text-davinci-003 by prompting \textit{Let's think step by step.}}\\Fact 1 states that climate is generally described in terms of temperature and moisture. Fact 2 states that humidity is a measure of moisture content of air. Therefore, the answer is (A) humidity.};

            \node(Davinci Fundamental Text)[rectangle, minimum width=1.1\textwidth, minimum height=0.04\textheight, fill=light-blue!45, xshift=0\textwidth, text width=1.35\textwidth, align=left, below of=Coordinate, yshift=-0.09\textheight]{\textbf{text-davinci-003 by prompting \textit{The fundamental of this question is:}}\\Climate is generally described in terms of temperature and moisture. Humidity is a measure of moisture content of air, so the correct answer is (A) humidity.};

            \node(Davinci Error Text)[rectangle, minimum width=1.1\textwidth, minimum height=0.04\textheight, fill=light-blue!60, xshift=0\textwidth, text width=1.35\textwidth, align=left, below of=Coordinate, yshift=-0.135\textheight]{\textbf{text-davinci-003 by prompting \textit{The common mistakes are:}}\\ (B) wind, (C) warmer air, (D) calibrated, (E) percentages, (F) matter, (G) color, and (H) SI units.};

           \node(Chain-of-Thought Text)[rectangle, minimum width=1.1\textwidth, minimum height=0.04\textheight, fill=light-green!30, xshift=0\textwidth, text width=1.35\textwidth, align=left, below of=Coordinate, yshift=-0.18\textheight]{\textbf{TeacherLM-7.1B by prompting \textit{Let's think step by step.}}\\Various physical properties describe the climate. One of them is humidity. The climate is generally described in terms of temperature and humidity. So, option humidity is the correct answer.};

            \node(Fundamental Text)[rectangle, minimum width=1.1\textwidth, minimum height=0.04\textheight, fill=light-green!45, xshift=0\textwidth, text width=1.35\textwidth, align=left, below of=Coordinate, yshift=-0.24\textheight]{\textbf{TeacherLM-7.1B by prompting \textit{The fundamental of this question is:}}\\Climate refers to the weather pattern in a long-term condition in a specific region. Several factors influence the climate of a specific region. Temperature, wind speed, humidity, precipitation, and solar radiation are the important significant factors that influence the climate of a specific region.};

            \node(Error Text)[rectangle, minimum width=1.1\textwidth, minimum height=0.04\textheight, fill=light-green!60, xshift=0\textwidth, text width=1.35\textwidth, align=left, below of=Coordinate, yshift=-0.305\textheight]{\textbf{TeacherLM-7.1B by prompting \textit{The common mistakes are:}}\\Do not think that climate is simply determined by temperature alone. Temperature, humidity, wind speed, and other factors also play an important role in the climate of a region. Do not consider that climate is not measured in humid regions. Climate is measured in terms of humidity also.};

        \end{tikzpicture}

    \end{subfigure}
    \caption{
    Example showing a common sense question, with its manual explanations from the original dataset and CoT, fundamentals, and common mistakes generated by text-davinci-003 and TeacherLM-7.1B-7.1B, from top to bottom.
    }
    \label{fig:example_output}
\end{figure*}
In this section, we select TeacherLM-7.1B as the teacher
for the following experiments in order to balance efficiency and performance. 
%We expand from three dimensions to verify the ability of TeacherLM.
Our experiment contains three dimensions to verify the ability of TeacherLM. The first dimension is the choice of different training modes, divided into multi-task training followed by zero-shot testing and single-task fine-tuning followed by testing in the corresponding task. The second dimension is the scaling of the model size. The third dimension is the diversity of the test tasks.

\subsection{Datasets}
We selected P3-Sense-3K for multi-task training. Moreover, for single-task fine-tuning, we aimed to compare TeacherLM-7.1B's enhancement effects with manually annotated rationales. Thus we selected (1) StrategyQA \cite{geva2021did}, a question-answering benchmark with implicit reasoning strategies. Due to the unavailability of the test dataset, the training dataset was was randomly redivided into training, validation, and test sets in an 8:1:1 ratio. (2) CREAK \cite{onoe2021creak}, a dataset for commonsense reasoning over entity knowledge. Since the test dataset lacks labels, we directly evaluated the dev dataset. (3) ECQA \cite{aggarwal2021explanations}, a dataset containing explanations for commonsenseQA.

\subsection{Models and training details}
We used the three models in the TeacherLM-7.1B series to augment each sample in the above four datasets. The augmented samples, as shown in Figure \ref{fig:example_output}, include five parts of information: question, answer, fundamentals, chain of thought, and common mistakes.
For more information on the above datasets, please refer to Table \ref{tab:dataset}.

In the multi-task training mode, we evaluated the benefit of the P3-Sense-3K dataset augmented by TeacherLM-7.1B on various models. The three parts of TeacherLM's explanations are concatenated into sequence with the answer. The control group consisted of the original P3-Sense-3k dataset, containing only question and answer pairs. We increased the student model size from 1.1B to 7.1B and set the learning rate for all experiments in the multi-task training mode to 2e-5 and the batch size to 256.

For the single-task fine-tuning, we select BLOOMZ-7.1B as the student model, which has been fine-tuned on xP3, a composite of supervised datasets in 46 languages with English and machine-translated prompts. In the single-task fine-tuning, we set the learning rate of all experiments to 6e-6 and the batch size to 64.

\subsection{Comparison with human and text-davinci-003}
To further validate model-generated explanations' quality, we include human annotation and text-davinci-003 as control groups in our experiments, where text-davinci-003 serves as the teacher and augments the StrategyQA, CREAK, and ECQA datasets in the same way as TeacherLM-7.1B.

Apart from training results in section~\ref{sec:training_results}, we found that manual explanations and text-davinci-003's augmentation both read smoothly but are inherently less detailed than TeacherLM-7.1B's. Nevertheless, text-davinci-003 sometimes reiterates the content of the question, rendering limited augmented information that could help the student model improve its abilities.

\subsection{Results}
\label{sec:training_results}
\textbf{Multi-task training}
In this experiment section, the model underwent extensive data training and thoroughly learned the rationales of data augmentation. As shown in Figure \ref{fig:comparison}, models of different sizes were able to bring significant benefits in general. The full experiment results can be seen in appendix \ref{appendix:student_result}.

\textbf{Single-task training}
In this section, we divided the experiment into two parts. First, we directly fine-tune BLOOMZ-7.1B for each task. In the StrategyQA and CREAK datasets, TeacherLM-7.1B performs beyond manual annotation and has a more robust augmentation ability than text-davinci-003 in the CREAK task.

Second, we train BLOOMZ-7.1B on the augmented P3-Sense-3K dataset, obtaining P3-Augmented-BLOOMZ-7.1B, and then fine-tune it on single tasks. The experiment shows that this further improved the accuracy of the task and, to some extent, solved the problem of insufficient data. In StrategyQA and CREAK, TeacherLM-7.1B also shows the ability to enhance data beyond human annotation and text-davinci-003.

In addition, we found that including manual rationales during training can harm some datasets, such as ECQA. Our model was able to alleviate some problems to some extent, but it could not completely solve them.

\section{Discussion \& Conclusion}
\label{sec:Discussion & Conclusion}
\textbf{Which part of data augmentation is the most important?}

It cannot be generalized to all datasets. There are unavoidable differences between datasets, which often lead to varying requirements for data augmentation content, as shown in Table \ref{tab:bloomz}.
% There are unavoidable differences between datasets, leading to varying data augmentation requirements.
Therefore, CoT, fundamentals, and common mistakes form a complementary relationship, and these three elements can be separated or combined in different orders to enhance samples. We conducted detailed experiments, as shown in Appendix \ref{appendix:student_result}. However, in general, CoT brings the most benefits. 

\textbf{How can data augmentation effects be maintained in the case of limited data?}

In addition to the TeacherLM-7.1B model, we also released the augmented P3-Sense-3K dataset. If the data is limited, multi-task training can first be performed on this dataset and then fine-tuned on the small dataset, as Table ~\ref{tab:bloomz} shows.

\textbf{What did the student model learn from the augmented data?}

Learn by analogy. Though many of the rationales generated by TeacherLM are not always correct, in this paper, no measures were taken to filter the augmented content, and the student model still produces significant benefits. The correctness of the augmented content is not the only important factor. The relevance of the content and the consistency of the reasoning logic are also significant \cite{wang2022towards}, which allows the student models to learn to think critically and truly enhance the model's generalization ability on unseen tasks.

\textbf{In comparison to human annotation and text-davinci-003, what are the characteristics and deficiencies of the generation content of TeacherLM-7.1B?}

As shown in Figure ~\ref{fig:example_output} and Appendix ~\ref{appendix:more_exmples}, TeacherLM-7.1B's explanations are generally more comprehensive and detailed than human annotations and text-davinci-003's explanations. However, it falls behind text-davinci-003 in solving mathematical problems, probably related to the model's size.

\bibliography{main}
\bibliographystyle{icml2023}

\newpage
\appendix
\onecolumn
\pdfoutput=1
\section{TeacherLMs}
\subsection{Training hyperparameters}
We show the training hyperparameters of TeacheLMs and the amount of training data at each stage in Table \ref{TeacherLM_parameters}.
\label{appendix:hyperparameters}
\begin{table}[htbp]
\caption{Training hyperparameter settings for each model size. TeacherLM-176B completed part of the training process and only trained on TeacherData-2M.}
\label{TeacherLM_parameters}
\begin{center}
\begin{tabular}{lcccccc}
\toprule
parameters & learning rate & batch size & tokens in 1st stage & tokens in 2nd stage &tokens in 3rd stage \\
\midrule
560M    & 3e-4 & 256 & 5B & 7.5B & 0.5B\\
1.1B & 1e-4 & 512 & 5B & 7.5B & 0.5B \\
3B    & 4e-5 & 512 & 5B &7.5B & 0.5B \\
7.1B    & 2e-5 & 768 & 0.5B & 8B & 1.5B  \\ 
176B  & 6e-5 & 1024 & / & 1B & /& \\
\bottomrule
\end{tabular}
\end{center}
\end{table}

\subsection{Full experimental results}
\label{appendix:result}
The evaluation results of TeacherLMs can be seen at Table \ref{tab:MMLU} and Table \ref{tab:MMLU_small}, where we show the MMLU individual task performance of TeacherLM-560M, TeacherLM-1.1B, TeacherLM-3B, FLAN-PALM-8B, TeacherLM-7.1B, TeacherLM-176B, GLM-130B, BLOOM-176B and Gopher-280B. Here, we report the “validation” set performance of individual tasks in MMLU.
\begin{table*}[t]
    \centering
        \caption{MMLU individual task performance of TeacherLM-7.1B, TacherLM-176B, GLM-130B, BLOOM-176B, and Gopher-280B. Furthermore, we denote ``college'' as ``CO'' and ``high school'' as ``HS''.}
        \label{tab:MMLU}
        \vskip 0.15in
        \begin{center}
        \begin{small}
        \begin{sc}
            \begin{tabular}{lccccc}
                \toprule
                task                                    & TeacherLM-7.1B & TeacherLM-176B & GLM-130B & BLOOM-176B & Gopher-280B \\
                \midrule
                abstract\_algebra                       & 30.00          & 32.00          & 24.00    & 24.00      & 25.00       \\
                anatomy                                 & 46.67          & 54.81          & 48.90    & 38.52      & 56.30       \\
                astronomy                               & 51.32          & 67.76          & 48.03    & 34.87      & 65.80       \\
                business\_ethics                        & 65.00          & 67.00          & 51.00    & 34.00      & 70.00       \\
                clinical\_knowledge                     & 57.74          & 62.26          & 48.68    & 35.85      & 67.20       \\
                co\_biology                        & 59.72          & 66.67          & 47.22    & 37.50      & 70.80       \\
                co\_chemistry                      & 42.00          & 42.00          & 34.00    & 19.00      & 45.00       \\
                co\_computer\_science              & 40.00          & 35.00          & 44.00    & 1.00       & 49.00       \\
                co\_mathematics                    & 34.00          & 35.00          & 27.00    & 31.00      & 37.00       \\
                co\_medicine                       & 50.29          & 58.96          & 43.35    & 28.90      & 60.10       \\
                co\_physics                        & 41.18          & 44.12          & 30.39    & 24.50      & 34.30       \\
                computer\_security                      & 64.00          & 76.00          & 61.00    & 40.00      & 65.00       \\
                conceptual\_physics                     & 51.49          & 51.06          & 38.72    & 31.49      & 49.40       \\
                econometrics                            & 35.09          & 44.74          & 26.32    & 26.32      & 43.00       \\
                electrical\_engineering                 & 53.10          & 60.00          & 45.52    & 32.41      & 60.00       \\
                elementary\_mathematics                 & 39.42          & 42.86          & 31.75    & 29.63      & 33.60       \\
                formal\_logic                           & 38.10          & 39.68          & 27.78    & 23.02      & 35.70       \\
                global\_facts                           & 29.00          & 32.00          & 35.00    & 23.00      & 38.00       \\
                hs\_biology                   & 62.90          & 70.97          & 51.29    & 27.42      & 71.30       \\
                hs\_chemistry                 & 43.35          & 50.74          & 34.98    & 27.09      & 47.80       \\
                hs\_computer\_science         & 62.00          & 64.00          & 53.00    & 30.00      & 54.00       \\
                hs\_european\_history         & 61.21          & 75.76          & 58.18    & 35.76      & 72.10       \\
                hs\_geography                 & 67.68          & 78.28          & 53.54    & 36.36      & 76.80       \\
                hs\_government\_and\_politics & 68.39          & 75.13          & 62.18    & 40.41      & 83.90       \\
                hs\_macroeconomics            & 55.64          & 62.82          & 42.56    & 30.77      & 65.10       \\
                hs\_mathematics               & 29.26          & 33.33          & 28.15    & 25.93      & 23.70       \\
                hs\_microeconomics            & 60.50          & 72.27          & 45.80    & 26.89      & 66.40       \\
                hs\_physics                   & 30.46          & 37.09          & 29.80    & 30.46      & 33.80       \\
                hs\_psychology                & 70.83          & 82.20          & 54.13    & 39.27      & 81.80       \\
                hs\_statistics                & 44.91          & 50.00          & 38.43    & 26.39      & 50.00       \\
                hs\_us\_history               & 54.90          & 66.18          & 58.33    & 40.69      & 78.90       \\
                hs\_world\_history            & 64.98          & 73.42          & 67.09    & 32.07      & 75.10       \\
                human\_aging                            & 56.95          & 67.26          & 45.29    & 32.29      & 66.40       \\
                human\_sexuality                        & 61.07          & 67.18          & 51.15    & 35.11      & 67.20       \\
                international\_law                      & 71.90          & 78.51          & 56.20    & 42.15      & 77.70       \\
                jurisprudence                           & 62.96          & 70.37          & 43.52    & 35.19      & 71.30       \\
                logical\_fallacies                      & 52.15          & 61.96          & 57.06    & 31.29      & 72.40       \\
                machine\_learning                       & 29.46          & 42.86          & 40.18    & 29.46      & 41.10       \\
                management                              & 75.73          & 73.79          & 56.31    & 27.18      & 77.70       \\
                marketing                               & 79.06          & 86.75          & 67.52    & 39.74      & 83.30       \\
                medical\_genetics                       & 57.00          & 71.00          & 48.00    & 45.00      & 69.00       \\
                miscellaneous                           & 60.54          & 71.14          & 61.18    & 40.23      & 75.70       \\
                moral\_disputes                         & 55.20          & 63.87          & 47.11    & 36.71      & 66.80       \\
                moral\_scenarios                        & 22.35          & 29.72          & 24.25    & 24.36      & 40.20       \\
                nutrition                               & 56.86          & 64.38          & 50.65    & 32.35      & 69.90       \\
                philosophy                              & 52.73          & 66.24          & 45.34    & 35.37      & 68.80       \\
                prehistory                              & 50.62          & 69.14          & 50.93    & 40.43      & 67.60       \\
                professional\_accounting                & 36.17          & 45.74          & 35.46    & 28.72      & 44.30       \\
                professional\_law                       & 34.16          & 42.31          & 37.94    & 29.53      & 44.50       \\
                professional\_medicine                  & 47.79          & 55.15          & 43.38    & 18.01      & 64.00       \\
                professional\_psychology                & 47.06          & 63.07          & 42.48    & 31.54      & 68.10       \\
                public\_relations                       & 66.36          & 64.55          & 55.46    & 33.64      & 71.80       \\
                security\_studies                       & 63.67          & 68.16          & 44.90    & 34.29      & 64.90       \\
                sociology                               & 70.15          & 81.09          & 51.74    & 31.84      & 84.10       \\
                us\_foreign\_policy                     & 70.00          & 78.00          & 61.00    & 46.00      & 81.00       \\
                virology                                & 42.17          & 46.99          & 39.16    & 28.31      & 47.00       \\
                world\_religions                        & 55.56          & 73.68          & 55.56    & 42.11      & 84.20       \\
                average                                 & 52.30          & 59.80          & 45.70    & 31.90      & 60.00      \\
                \bottomrule
                \end{tabular}
        \end{sc}
        \end{small}
        \end{center}
        \vskip -0.1in
        \end{table*}
\begin{table*}[t]
    \centering
        \caption{MMLU individual task performance of TeacherLM-560M, TacherLM-1.1B, TeacherLM-3B, Flan-PaLM-8B. Furthermore, we denote ``college'' as ``CO'' and ``high school'' as ``HS''.}
        \label{tab:MMLU_small}
        \vskip 0.15in
        \begin{center}
        \begin{small}
        \begin{sc}
                 \begin{tabular}{lcccc}
                    \toprule
                    task                                    & TeacherLM-560M & TeacherLM-1.1B & TeacherLM-3B & Flan-PaLM -8B \\
                    \midrule
                    abstract\_algebra                       & 29.00          & 33.00          & 32.00        & 36.40         \\
                    anatomy                                 & 42.96          & 39.26          & 40.74        & 42.90         \\
                    astronomy                               & 44.08          & 42.11          & 46.71        & 43.80         \\
                    business\_ethics                        & 30.00          & 44.00          & 51.00        & 36.40         \\
                    clinical\_knowledge                     & 47.92          & 45.66          & 53.21        & 48.30         \\
                    co\_biology                        & 36.11          & 35.42          & 47.92        & 56.20         \\
                    co\_chemistry                      & 37.00          & 31.00          & 42.00        & 25.00         \\
                    co\_computer\_science              & 35.00          & 40.00          & 30.00        & 54.50         \\
                    co\_mathematics                    & 30.00          & 32.00          & 41.00        & 18.20         \\
                    co\_medicine                       & 47.40          & 40.46          & 47.40        & 50.00         \\
                    co\_physics                        & 32.35          & 30.39          & 37.25        & 45.50         \\
                    computer\_security                      & 42.00          & 49.00          & 58.00        & 72.70         \\
                    conceptual\_physics                     & 37.02          & 41.28          & 44.68        & 38.50         \\
                    econometrics                            & 28.07          & 33.33          & 26.32        & 33.30         \\
                    electrical\_engineering                 & 46.20          & 46.90          & 49.66        & 37.50         \\
                    elementary\_mathematics                 & 29.89          & 34.13          & 34.66        & 34.10         \\
                    formal\_logic                           & 29.37          & 26.98          & 32.54        & 21.40         \\
                    global\_facts                           & 24.00          & 30.00          & 30.00        & 30.00         \\
                    hs\_biology                   & 40.32          & 50.65          & 57.42        & 50.00         \\
                    hs\_chemistry                 & 39.90          & 39.41          & 43.35        & 18.20         \\
                    hs\_computer\_science         & 40.00          & 38.00          & 47.00        & 44.40         \\
                    hs\_european\_history         & 42.42          & 41.21          & 52.73        & 72.20         \\
                    hs\_geography                 & 44.95          & 51.52          & 56.57        & 68.20         \\
                    hs\_government\_and\_politics & 40.93          & 47.67          & 53.37        & 57.10         \\
                    hs\_macroeconomics            & 38.46          & 41.28          & 53.08        & 44.20         \\
                    hs\_mathematics               & 28.52          & 27.04          & 26.67        & 17.20         \\
                    hs\_microeconomics            & 41.60          & 50.00          & 60.08        & 57.70         \\
                    hs\_physics                   & 29.14          & 29.14          & 29.14        & 17.60         \\
                    hs\_psychology                & 44.22          & 55.23          & 63.85        & 68.30         \\
                    hs\_statistics                & 35.65          & 34.26          & 41.20        & 39.10         \\
                    hs\_us\_history               & 39.22          & 40.69          & 47.55        & 72.70         \\
                    hs\_world\_history            & 40.51          & 45.15          & 53.59        & 61.50         \\
                    human\_aging                            & 39.46          & 40.81          & 55.61        & 52.20         \\
                    human\_sexuality                        & 44.27          & 43.51          & 48.09        & 66.70         \\
                    international\_law                      & 58.68          & 65.29          & 62.81        & 76.90         \\
                    jurisprudence                           & 44.44          & 49.07          & 50.00        & 72.70         \\
                    logical\_fallacies                      & 32.52          & 34.97          & 46.01        & 61.10         \\
                    machine\_learning                       & 28.57          & 27.68          & 36.61        & 45.50         \\
                    management                              & 44.66          & 63.11          & 61.17        & 81.80         \\
                    marketing                               & 50.85          & 63.25          & 67.95        & 72.00         \\
                    medical\_genetics                       & 42.00          & 39.00          & 48.00        & 63.60         \\
                    miscellaneous                           & 37.42          & 43.93          & 54.92        & 68.60         \\
                    moral\_disputes                         & 45.95          & 45.09          & 52.02        & 39.50         \\
                    moral\_scenarios                        & 27.15          & 23.58          & 24.69        & 25.00         \\
                    nutrition                               & 45.42          & 47.06          & 50.33        & 57.60         \\
                    philosophy                              & 35.05          & 44.37          & 47.59        & 61.80         \\
                    prehistory                              & 37.04          & 38.27          & 42.90        & 45.70         \\
                    professional\_accounting                & 35.46          & 32.98          & 35.82        & 35.50         \\
                    professional\_law                       & 29.40          & 32.40          & 33.64        & 32.40         \\
                    professional\_medicine                  & 29.78          & 33.82          & 42.65        & 51.60         \\
                    professional\_psychology                & 33.82          & 40.36          & 43.46        & 46.40         \\
                    public\_relations                       & 32.73          & 42.73          & 50.91        & 50.00         \\
                    security\_studies                       & 44.08          & 47.76          & 54.69        & 40.70         \\
                    sociology                               & 54.72          & 62.69          & 65.67        & 72.70         \\
                    us\_foreign\_policy                     & 52.00          & 57.00          & 67.00        & 63.60         \\
                    virology                                & 33.13          & 36.14          & 40.96        & 44.40         \\
                    world\_religions                        & 35.67          & 38.01          & 50.29        & 68.40         \\
                    average                                 & 38.39          & 41.38          & 46.74        & 49.29        \\
                    \bottomrule
                    \end{tabular}
        \end{sc}
        \end{small}
        \end{center}
        \vskip -0.1in
        \end{table*}
\section{Complete Benchmarks for All Augmented Trained Models}
We evaluate all the augmented trained models on four benchmarks, MMLU, ECQA, CREAK and StrategyQA. We show the results in Table \ref{tab:four_datasets}. We evaluate ten kinds of checkpoints, including original pretrained models, models trained with original P3 dataset, models trained with P3-Sense-3K, and models trained with seven kinds of augmented dataset.
\label{appendix:student_result}
\begin{table*}[t]
    \caption{
    Comparing different benchmarks of the original six models we use and the scores of each model after training with P3 dataset, P3-Sense-3K, and seven augmented datasets. PRETRAINED in the figure represents the original models; ORIGIN represents models trained on origin P3 dataset; SENSE represents models trained on P3-Sense-3K; The rest represent models trained on augmented P3 dataset in different prompt format. C represents the CoT field; M represents the common mistakes field; F represents the fundamental field.
    }
    \label{tab:four_datasets}
    \vskip 0.15in
    \begin{center}
    \begin{small}
    \begin{sc}
        \begin{subtable}{1\textwidth}
            \begin{tabular}{lccccccccc}
            \toprule
            \textbf{MMLU } & BL-1B1 & BL-3B & BL-7B1 & BZ-1B1 & BZ-3B & BZ-7B1 & OPT-1B3 & OPT-2B7 & OPT-6B7 \\
            \midrule
            Pretrained & 23.1      & 25.1     & 24.7      & 23.7       & 35.2      & 39.6       & 24.8    & 22.2    & 24.9    \\
            Origin     & 25.8      & 26.3     & 32.8      & 24.8       & 28.9      & 32.2       & 27.8    & 28.3    & 31.0    \\
            Sense      & 31.0      & 36.1     & 39.2      & 32.2       & 38.6      & 41.3       & 33.1    & 37.0    & 39.1    \\
            C          & 32.9      & 38.1     & 42.6      & 32.8       & 40.1      & 44.0       & 35.8    & 38.4    & 42.2    \\
            C\_M       & 32.5      & 37.4     & 42.3      & 33.1       & 40.2      & 44.1       & 34.9    & 38.4    & 41.3    \\
            C\_M\_F    & 33.0      & 38.0     & 43.0      & 33.5       & 40.3      & 44.7       & 34.3    & 39.4    & 42.7    \\
            shuffle    & 33.1      & 38.2     & 41.0      & 32.9       & 40.1      & 44.1       & 35.3    & 38.2    & 41.6    \\
            M          & 31.9      & 37.3     & 41.5      & 32.8       & 39.0      & 43.1       & 34.8    & 38.1    & 39.9    \\
            F          & 31.7      & 37.0     & 42.0      & 32.7       & 39.7      & 43.5       & 33.2    & 37.9    & 40.9    \\
            F\_C\_M    & 32.5      & 38.0     & 43.2      & 32.6       & 40.2      & 44.5       & 33.4    & 39.5    & 42.2   \\
            \bottomrule
            \end{tabular}
        \end{subtable}
        
        ~\\

        ~\\
        \begin{subtable}{1\textwidth}
            \begin{tabular}{lccccccccc}
                \toprule
                \textbf{ECQA } & BL-1B1 & BL-3B & BL-7B1 & BZ-1B1 & BZ-3B & BZ-7B1 & OPT-1B3 & OPT-2B7 & OPT-6B7 \\
                \midrule
                Pretrained & 20.6      & 19.3     & 21.2      & 20.2       & 62.8      & 74.3       & 21.2    & 20.4    & 20.8    \\
                Origin     & 22.1      & 46.0     & 61.3      & 24.8       & 54.5      & 64.3       & 27.6    & 48.5    & 64.9    \\
                Sense      & 50.0      & 63.4     & 70.0      & 32.2       & 76.0      & 78.2       & 59.5    & 69.6    & 71.9    \\
                C          & 51.5      & 65.3     & 71.9      & 32.8       & 75.6      & 80.1       & 59.5    & 70.4    & 68.8    \\
                C\_M       & 53.1      & 64.7     & 71.7      & 33.1       & 76.4      & 79.5       & 58.8    & 70.2    & 71.9    \\
                C\_M\_F    & 49.4      & 64.8     & 70.9      & 33.5       & 74.5      & 78.2       & 51.9    & 69.7    & 71.7    \\
                shuffle    & 55.7      & 67.9     & 71.3      & 32.9       & 75.8      & 78.2       & 58.1    & 71.6    & 72.9    \\
                M          & 51.8      & 65.5     & 71.4      & 32.8       & 76.6      & 78.6       & 62.1    & 70.2    & 71.6    \\
                F          & 49.1      & 63.4     & 70.1      & 32.7       & 75.0      & 78.6       & 49.7    & 69.3    & 70.3    \\
                F\_C\_M    & 46.5      & 64.8     & 70.6      & 32.6       & 75.3      & 78.2       & 50.0    & 68.4    & 71.8   \\
                \bottomrule
            \end{tabular}
        \end{subtable}
        
        ~\\

        ~\\
        \begin{subtable}{1\textwidth}
            \begin{tabular}{lccccccccc}
                    \toprule
                    \textbf{CREAK } & BL-1B1 & BL-3B & BL-7B1 & BZ-1B1 & BZ-3B & BZ-7B1 & OPT-1B3 & OPT-2B7 & OPT-6B7 \\
                    \midrule
                Pretrained & 50.6      & 49.6     & 49.6      & 50.4       & 49.6      & 49.6       & 49.7    & 53.7    & 64.6    \\
                Origin     & 51.3      & 54.6     & 51.6      & 51.3       & 51.4      & 51.1       & 52.2    & 51.2    & 51.9    \\
                Sense      & 59.2      & 68.2     & 67.9      & 58.8       & 66.7      & 71.3       & 67.1    & 70.2    & 73.6    \\
                C          & 61.4      & 66.8     & 71.3      & 63.7       & 67.0      & 70.6       & 68.8    & 73.2    & 74.4    \\
                C\_M       & 62.4      & 68.8     & 71.8      & 63.7       & 68.8      & 72.2       & 68.3    & 72.9    & 75.7    \\
                C\_M\_F    & 61.9      & 70.2     & 70.5      & 63.9       & 69.0      & 70.8       & 65.9    & 73.5    & 75.0    \\
                shuffle    & 60.0      & 68.6     & 72.0      & 63.7       & 69.0      & 70.8       & 67.9    & 72.6    & 75.0    \\
                M          & 60.7      & 67.0     & 73.2      & 62.7       & 66.7      & 71.7       & 66.4    & 71.9    & 75.2    \\
                F          & 57.4      & 66.9     & 70.8      & 60.7       & 67.6      & 71.1       & 69.1    & 72.6    & 73.8    \\
                F\_C\_M    & 64.1      & 69.3     & 70.4      & 62.9       & 68.5      & 70.9       & 65.0    & 73.2    & 75.9   \\
                \bottomrule
            \end{tabular}
        \end{subtable}
        
        ~\\

        ~\\
        \begin{subtable}{1\textwidth}
            \begin{tabular}{lccccccccc}
                \toprule
                \textbf{StrategyQA }   & BL-1B1 & BL-3B & BL-7B1 & BZ-1B1 & BZ-3B & BZ-7B1 & OPT-1B3 & OPT-2B7 & OPT-6B7 \\
                \midrule
                Pretrained & 55.3      & 53.5     & 52.2      & 50.0       & 52.2      & 47.4       & 53.5    & 63.6    & 71.5    \\
                Origin     & 57.0      & 55.7     & 60.5      & 56.6       & 52.6      & 63.2       & 57.9    & 54.4    & 63.2    \\
                Sense      & 57.5      & 69.7     & 70.6      & 61.8       & 71.5      & 73.7       & 62.3    & 65.8    & 71.9    \\
                C          & 58.8      & 68.9     & 72.4      & 64.0       & 70.6      & 73.2       & 65.4    & 66.7    & 74.6    \\
                C\_M       & 59.6      & 70.2     & 69.7      & 62.7       & 71.9      & 73.7       & 65.8    & 66.7    & 73.7    \\
                C\_M\_F    & 58.8      & 65.4     & 72.4      & 61.8       & 69.3      & 75.0       & 62.7    & 65.4    & 72.4    \\
                shuffle    & 62.3      & 71.5     & 71.9      & 61.8       & 71.1      & 73.7       & 64.0    & 65.4    & 73.2    \\
                M          & 61.8      & 66.2     & 74.6      & 64.5       & 72.4      & 71.5       & 66.2    & 68.4    & 74.1    \\
                F          & 60.1      & 68.0     & 71.9      & 63.6       & 70.2      & 75.0       & 63.2    & 68.9    & 70.6    \\
                F\_C\_M    & 60.1      & 65.8     & 71.1      & 64.0       & 70.2      & 72.8       & 60.5    & 66.2    & 71.5    \\
                \bottomrule
            \end{tabular}
        \end{subtable}
    \end{sc}
    \end{small}
    \end{center}
    \vskip -0.1in
    \end{table*}
\section{More Comparison Examples}
Figure \ref{fig:example_output} shows TeacherLM-7.1B's explanations for a Physics problem. Apart from Physics, TeacherLM-7.1B also has a sound ability to analyze History and Geography problems, which we show in Figure \ref{fig:more_example}.
\definecolor{light-green}{HTML}{99E2B4}
\definecolor{light-blue}{HTML}{CAF0F8}
\definecolor{light-gray}{HTML}{EDEDE9}
\definecolor{grey}{HTML}{D6CCC2}

\begin{figure*}[ht]
    \begin{subfigure}{0.74\textwidth}\footnotesize
        \begin{tikzpicture}

            \node(bounding box)[rectangle, minimum width=1\textwidth, minimum height=0\textheight, draw=none, dashed]{};

            \node(Input Text)[rectangle, minimum width=1.1\textwidth, minimum height=0.04\textheight, fill=grey!10, xshift=0\textwidth, text width=1.35\textwidth, align=left, yshift=0.09\textheight]{\textbf{Question:} Determine the correctness of the following sentence. Sam Cooke became a member of the Rock \& Roll Hall of Fame. \\ (A) True (B) False \\ \textbf{Answer:} The answer is (A).};

            \node(Coordinate)[rectangle, rounded corners=0.0\textwidth, minimum width=0.0\textwidth, minimum height=0.0\textheight, below of=input box, yshift=-0.025\textheight]{};

            %! 这是全局的坐标，不显示任何字样

            \node(Manual Text)[rectangle, minimum width=1.1\textwidth, minimum height=0.05\textheight, fill=grey!15, xshift=0\textwidth, text width=1.35\textwidth, align=left, below of=Coordinate, yshift=+0.005 \textheight]{\textbf{Manual Explanations}\\I know this to be true.};

            \node(Davinci Chain-of-Thought Text)[rectangle, minimum width=1.1\textwidth, minimum height=0.04\textheight, fill=light-blue!30, xshift=0\textwidth, text width=1.35\textwidth, align=left, below of=Coordinate, yshift=-0.04\textheight]{\textbf{text-davinci-003 by prompting \textit{Let's think step by step.}}\\ Sam Cooke was an American singer, songwriter, and entrepreneur. He was one of the most influential musical artists of the 20th century. He was inducted into the Rock and Roll Hall of Fame in 1986. Therefore, the statement is true.};

            \node(Davinci Fundamental Text)[rectangle, minimum width=1.1\textwidth, minimum height=0.04\textheight, fill=light-blue!45, xshift=0\textwidth, text width=1.35\textwidth, align=left, below of=Coordinate, yshift=-0.08\textheight]{\textbf{text-davinci-003 by prompting \textit{The fundamental of this question is:}}\\True. Sam Cooke was inducted into the Rock and Roll Hall of Fame in 1986.};

            \node(Davinci Error Text)[rectangle, minimum width=1.1\textwidth, minimum height=0.04\textheight, fill=light-blue!60, xshift=0\textwidth, text width=1.35\textwidth, align=left, below of=Coordinate, yshift=-0.12\textheight]{\textbf{text-davinci-003 by prompting \textit{The common mistakes are:}}\\ Sam Cooke became a member of the Rock and Roll Hall of Fame.};

           \node(Chain-of-Thought Text)[rectangle, minimum width=1.1\textwidth, minimum height=0.04\textheight, fill=light-green!30, xshift=0\textwidth, text width=1.35\textwidth, align=left, below of=Coordinate, yshift=-0.16 \textheight]{\textbf{TeacherLM-7.1B by prompting \textit{Let's think step by step.}}\\Step 1: The early 1970s saw the emergence of Rock \& Roll as the leading music form in the United States. Step 2: Sam Cooke is known as the King of Soul. He became a Rock \& Roll Hall of Fame member in 1995.};

            \node(Fundamental Text)[rectangle, minimum width=1.1\textwidth, minimum height=0.04\textheight, fill=light-green!45, xshift=0\textwidth, text width=1.35\textwidth, align=left, below of=Coordinate, yshift=-0.2357 \textheight]{\textbf{TeacherLM-7.1B by prompting \textit{The fundamental of this question is:}}\\The United States established several organizations to recognize excellence in achievements. These organizations are the National Foundation for the Advancement of Arts and Sciences and the Rock and Roll Hall of Fame. In 1972, the Country Music Association introduced the first-ever Rock and Roll Hall of Fame. A rock music awards show named the BBC Radio 1 Big Weekend also emerged. This award show had several celebrities competing for the top honor: the Rock and Roll Hall of Fame. The first inductee was the band, the Beach Boys. The recording industry's successes maintain the honor.};

            \node(Error Text)[rectangle, minimum width=1.1\textwidth, minimum height=0.04\textheight, fill=light-green!60, xshift=0\textwidth, text width=1.35\textwidth, align=left, below of=Coordinate, yshift=-0.293 \textheight]{\textbf{TeacherLM-7.1B by prompting \textit{The common mistakes are:}}\\Rock \& Roll songs emerged in the 1950s and 1960s. Do not forget that Ray Charles and Sam Cooke are both in the Rock \& Roll Hall of Fame.};

        \end{tikzpicture}

        \begin{tikzpicture}

            \node(Coordinate)[rectangle, rounded corners=0.0\textwidth, minimum width=0.0\textwidth, minimum height=0.05\textheight, below of=input box, yshift=-0.025\textheight]{};

        \end{tikzpicture}

        \begin{tikzpicture}

            \node(bounding box)[rectangle, minimum width=1\textwidth, minimum height=0\textheight, draw=none, dashed]{};

            \node(Input Text)[rectangle, minimum width=1.1\textwidth, minimum height=0.04\textheight, fill=grey!10, xshift=0\textwidth, text width=1.35\textwidth, align=left, yshift=0.09\textheight]{\textbf{Question:} Determine the correctness of the following sentence. Turmeric grows in the sub-Asian region of the world. \\ (A) True (B) False \\ \textbf{Answer:} The answer is (A).};

            \node(Coordinate)[rectangle, rounded corners=0.0\textwidth, minimum width=0.0\textwidth, minimum height=0.0\textheight, below of=input box, yshift=-0.025\textheight]{};

            %! 这是全局的坐标，不显示任何字样

            \node(Manual Text)[rectangle, minimum width=1.1\textwidth, minimum height=0.05\textheight, fill=grey!15, xshift=0\textwidth, text width=1.35\textwidth, align=left, below of=Coordinate, yshift=+0.005 \textheight]{\textbf{Manual Explanations}\\Because they are facts.};

            \node(Davinci Chain-of-Thought Text)[rectangle, minimum width=1.1\textwidth, minimum height=0.04\textheight, fill=light-blue!30, xshift=0\textwidth, text width=1.35\textwidth, align=left, below of=Coordinate, yshift=-0.04\textheight]{\textbf{text-davinci-003 by prompting \textit{Let's think step by step.}}\\ Turmeric is a plant that is native to South Asia. Therefore, it grows in the sub-Asian region of the world. This means that the statement is true.};

            \node(Davinci Fundamental Text)[rectangle, minimum width=1.1\textwidth, minimum height=0.04\textheight, fill=light-blue!45, xshift=0\textwidth, text width=1.35\textwidth, align=left, below of=Coordinate, yshift=-0.084\textheight]{\textbf{text-davinci-003 by prompting \textit{The fundamental of this question is:}}Turmeric is a plant native to the subcontinent of India, which is part of the Asian region of the world. Therefore, the statement is true.\\};

            \node(Davinci Error Text)[rectangle, minimum width=1.1\textwidth, minimum height=0.041\textheight, fill=light-blue!60, xshift=0\textwidth, text width=1.35\textwidth, align=left, below of=Coordinate, yshift=-0.123\textheight]{\textbf{text-davinci-003 by prompting \textit{The common mistakes are:}}\\ True.};

           \node(Chain-of-Thought Text)[rectangle, minimum width=1.1\textwidth, minimum height=0.04\textheight, fill=light-green!30, xshift=0\textwidth, text width=1.35\textwidth, align=left, below of=Coordinate, yshift=-0.177 \textheight]{\textbf{TeacherLM-7.1B by prompting \textit{Let's think step by step.}}\\ Step 1: The botanical name of Turmeric is Curcuma Longa. The root of this plant is used for many purposes, like its anticancerous, analgesic, and antimicrobial properties. Step 2: Turmeric is also known as Indian ginger because the herbaceous root vegetable is native to south Asia, also called sub-Asian. So the statement is true.};

            \node(Fundamental Text)[rectangle, minimum width=1.1\textwidth, minimum height=0.04\textheight, fill=light-green!45, xshift=0\textwidth, text width=1.35\textwidth, align=left, below of=Coordinate, yshift=-0.2565 \textheight]{\textbf{TeacherLM-7.1B by prompting \textit{The fundamental of this question is:}}\\India has a tropical monsoon climate with four seasons throughout the year, a cool season from January to February, a summer season from March to May, a southwest monsoon rainy season from June to September, and a northeast monsoon period from October to December. The temperature is as low as 150C in the north and as high as 27C in the south. Almost all year is frost-free, crops can be grown throughout the year, and the heat resources are abundant. In this way, very many tropical plants can survive in India.};

            \node(Error Text)[rectangle, minimum width=1.1\textwidth, minimum height=0.04\textheight, fill=light-green!60, xshift=0\textwidth, text width=1.35\textwidth, align=left, below of=Coordinate, yshift=-0.328 \textheight]{\textbf{TeacherLM-7.1B by prompting \textit{The common mistakes are:}}\\Make sure to distinguish the term sub-Asia from Asia. Although sub-Asia is part of Asia, people generally refer to  India as sub-Asia.};

        \end{tikzpicture}
    \end{subfigure}
    \caption{
    More examples generated by TeacherLM-7.1B and text-davinci-003, with its manual explanations from the original dataset and CoT, fundamentals, and common mistakes generated by text-davinci-003 and TeacherLM-7.1B, from top to bottom.
    }
    \label{fig:more_example}
\end{figure*}
\label{appendix:more_exmples}
\pdfoutput=1
\end{document}